\pdfoutput=1

\documentclass[11pt]{article}

\usepackage[]{acl}
\usepackage{times}
\usepackage{latexsym}

\usepackage[T1]{fontenc}

\usepackage[utf8]{inputenc}
\usepackage{booktabs}
\usepackage{graphicx}
\usepackage{makecell}
\usepackage[font=small,labelfont=bf]{caption}
\usepackage{amsmath}
\usepackage{amssymb} 
\usepackage{multirow}
\usepackage{hyperref}
\usepackage{tabularx}
\usepackage{fancyvrb}  
\usepackage{listings}
\usepackage{tcolorbox}
\usepackage{enumitem}
\usepackage{soul, color}
\usepackage{arydshln}
\usepackage{listings}
\usepackage{xcolor}
\usepackage{colortbl}
\usepackage{float}
\usepackage{subcaption}
\definecolor{participantblue}{HTML}{003366}
\definecolor{modelgreen}{HTML}{006600}
\usepackage{url}

\usepackage{microtype}

\title{
An Offline Mobile Conversational Agent for Mental Health Support: Learning from Emotional Dialogues and Psychological Texts with Student-Centered Evaluation}

\author{
  \textbf{Vimaleswar A\textsuperscript{}}\textsuperscript{$\Delta$},
  \textbf{Prabhu Nandan Sahu\textsuperscript{}},
  \textbf{Nilesh Kumar Sahu\textsuperscript{}},
  \textbf{Haroon R. Lone\textsuperscript{}}
\\
\\
  \textsuperscript{}Department of Electrical Engineering and Computer Science, Indian Institute of \\ Science Education and Research Bhopal
\\
}

\begin{document}
\maketitle

\begingroup
  \renewcommand\thefootnote{$\Delta$}\footnotetext{Corresponding author: vimaleswar20@iiserb.ac.in}
\endgroup
\setcounter{footnote}{0}
\renewcommand\thefootnote{\arabic{footnote}}%

\begin{abstract}

Mental health plays a crucial role in the overall well-being of an individual. In recent years, digital platforms have increasingly been used to expand mental health and emotional support. However, there are persistent challenges related to limited user accessibility, internet connectivity, and data privacy, which highlight the need for an offline, smartphone-based solutions. To address these challenges, we propose \textbf{EmoSApp (Emotional Support App)}\footnote{Our data, app, models are available at \url{https://huggingface.co/SIR-Lab/EmoSApp}}: an entirely offline, smartphone-based conversational app designed to provide mental health and emotional support. EmoSApp leverages a language model, specifically the LLaMA-3.2-1B-Instruct, which is fine-tuned and quantized on a custom-curated ``Knowledge Dataset'' comprising 14,582 mental health QA pairs along with multi-turn conversational data, enabling robust domain expertise and fully on-device inference on resource-constrained smartphones.

Through qualitative evaluation with students and mental health professionals, we demonstrate that EmoSApp has the ability to respond coherently and empathetically, provide relevant suggestions to user’s mental health problems, and maintain interactive dialogue. Additionally, quantitative evaluations on nine commonsense and reasoning benchmarks, along with two mental health specific datasets, demonstrate EmoSApp's effectiveness in low-resource settings. By prioritizing on-device deployment and specialized domain-specific adaptation, EmoSApp serves as a blueprint for future innovations in portable, secure, and highly tailored AI-driven mental health support.\footnote{\bf This is a preprint of a paper accepted at IJCNLP–AACL 2025. The final authenticated version will appear in the conference proceedings.}


\end{abstract}

\begin{figure}[h]
    \centering
    \includegraphics[width=0.9\linewidth]{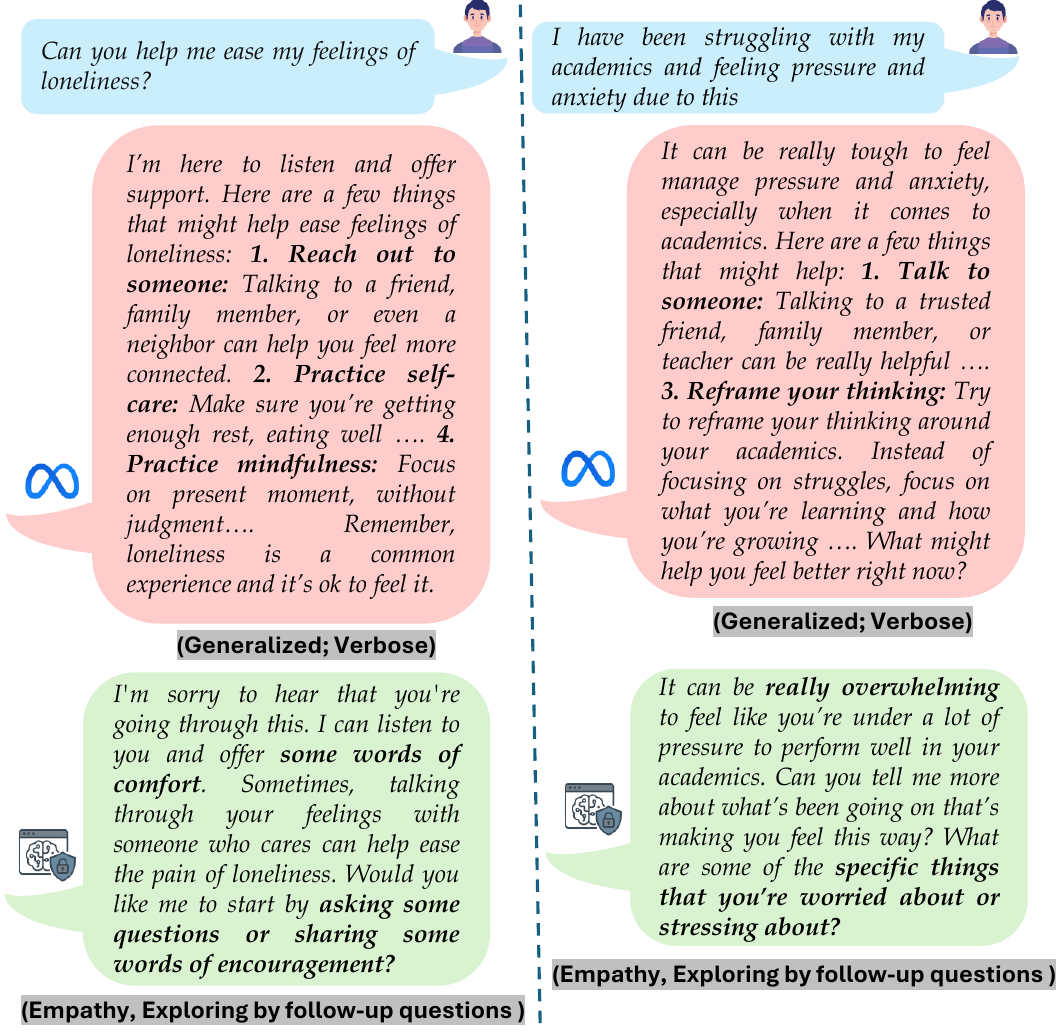}
    \caption{Example responses generated by LLaMA-3.2-1B-Instruct and our proposed EmoSApp. LLaMA model (pink response box) often give generalized and verbose responses, resulting in an AI-generated format. 
        In contrast, EmoSApp (green response box) demonstrates stronger empathy and deeper conversational exploration, effectively providing emotional and mental health support to users.}
    \label{fig:egconv}
\end{figure}
\vspace{0.5pt}

\section{Introduction}

Recent advancements in digital platforms and Natural Language Processing (NLP), particularly with the emergence of Large Language Models (LLMs) \cite{openai2024gpt4technicalreport, touvron2023llama2openfoundation, yang2024qwen2technicalreport}, have enabled the rise of conversational Artificial Intelligence (AI) for mental health support~\cite{LIU2022100495}, addressing several traditional barriers to care \cite{khosravi2024factors}. These technologies offer potential advantages, including ubiquitous availability, reduced costs, and scalability. Despite these advancements, persistent challenges related to user accessibility, the need for constant internet connectivity, data privacy, and high computational requirements limit the effectiveness of existing solutions \cite{coelho2025promoting, alhammad2024patients}.

Most existing conversational AI or chatbots for emotional and mental health support depends mainly on a computer-based design that requires continuous internet connectivity and transmit highly sensitive user data to external servers, thereby  amplifying privacy risks \cite{zhang2023adoption}. As depicted in Figure \ref{fig:egconv}, such chatbots often tend to be generalized, verbose and formulaic, often lacking the nuanced, context-specific guidance that real-world users require \cite{kang2024can, loh2023harnessing}. Furthermore, the computational demands of LLMs have precluded their deployment directly on resource-constrained devices, such as smartphones.

To address these gaps, we present {\bf EmoSApp}\footnote{\url{https://huggingface.co/SIR-Lab/EmoSApp}}, a fully offline, smartphone-based conversational AI app that designed to provide mental health and emotional support, particularly to students. Our approach leverages recent advances in model fine-tuning and quantization techniques, to enable on-device inference with LLMs on resource-constrained smartphone, specifically using Torchtune\footnote{\url{https://github.com/pytorch/torchtune}} and Executorch\footnote{\url{https://github.com/pytorch/executorch}} library. To our knowledge, we are the first to use Executorch and Torchtune in the mental health domain. By fine-tuning the LLaMA-3.2-1B-Instruct model\footnote{\url{https://huggingface.co/meta-llama/Llama-3.2-1B-Instruct}} on our custom-curated dataset from various psychology and mental health textbooks, comprising 14,582 mental health question-answer pairs, along with two multi-turn conversational corpora, we equip EmoSApp with domain-specific knowledge while maintaining practical computational efficiency for smartphone deployment.

The on-device nature of {\bf EmoSApp} directly addresses several critical limitations of existing approaches: (i) By running entirely offline, it eliminates internet-connectivity barriers, making it easily accessible to a broader range of smartphone users. (ii) By keeping all conversation data logs locally on the user's device, it enhances privacy protections for sensitive mental health conversations. (iii) By utilizing quantized models, it delivers responsive performance on modern smartphones with as little as 6GB of system RAM. (iv) Despite quantization, our qualitative evaluations show that EmoSApp maintains coherent, empathetic, and contextually relevant interactions that address users’ mental health concerns. By combining specialized domain adaptation through fine-tuning with efficient smartphone deployment, {\bf EmoSApp} represents a significant step towards making AI-driven emotional support for the student population more accessible, private, and contextually appropriate.

Our contributions are summarized in four folds:
\begin{itemize}
    \item We present \textbf{EmoSApp}, a fully functional offline emotional and mental health support app tailored for students, by fine-tuning and quantizing LLaMA-3.2-1B-Instruct optimized for resource-constrained smartphones.
    \item We propose a domain specialization approach via multi-source dataset fusion, integrating ESConv, ServeForEmo, and a custom-curated knowledge dataset derived from psychological and mental health textbooks. 
    \item We explore and compare multiple fine-tuning and quantization strategies (Full, LoRA + PTQ, QAT-LoRA) to balance model performance and on-device efficiency.
    \item Extensive quantitative and qualitative human evaluations demonstrate that EmoSApp excels in providing coherent, empathetic, and contextually appropriate support to students.
\end{itemize}

\section{Related Work}
The advancements and adaptability of Large Language Models (LLMs) have encouraged researchers to explore their use in mental health applications \cite{liu-etal-2021-towards}. LLMs have been applied to diverse tasks, such as predicting mental health conditions \cite{xu2024mental}, summarizing mental health conversations \cite{sahu2025leveraging} and offering support through conversational agents \cite{ye-etal-2025-sweetiechat}. With their extensive training and large parameter counts, models like ChatGPT, LLaMA, Gemini, and Claude have shown the potential to act as virtual counselors or mental health advisors \cite{song2024typing, malgaroli2025large}. As a result, some users have started exploring these models for mental health support \cite{song2024typing}. However, privacy remains a significant concern when using online LLM-based systems for sensitive and general tasks \cite{chandra2024lived, mandal2025towards}. To address this, researchers have begun developing lightweight LLMs that can run locally on personal devices \cite{mobilellm, thawakar2024mobillama}. However, running high-parameter LLMs locally requires powerful and expensive hardware. While these smaller models make on-device inference more feasible, the reduction in model size often results in diminished reasoning capacity and knowledge retention compared to larger models~\cite{deshpande2023honey}. Moreover, these models have primarily been optimized for general-purpose language understanding rather than for specialized domains such as mental health support.

\noindent \textbf{Language models and Mental Health:} Language models often lack empathy and do not follow clear support strategies when responding to users with emotional concerns. To address this, \cite{liu-etal-2021-towards} created a handcrafted dataset called Emotional Support Conversation (ESConv), which includes common support strategies people use when helping others. They fine-tuned models like DialoGPT and BlenderBot using this dataset. The fine-tuned models showed improved performance compared to the base versions by providing structured and strategy-driven emotional support. However, constructing such datasets at scale remains challenging due to the heavy reliance on human annotation.

Recent work has turned LLMs to scale up data generation. For instance, \cite{zheng-etal-2023-augesc} fine-tuned the 6B GPT-J model on ESConv to create a larger dataset called AugESC. However, these augmented datasets do not explicitly label or control the support strategies used, which limits interpretability. To address this limitation, \cite{ye-etal-2025-sweetiechat} proposed a multi-agent framework using three LLMs: a help seeker, a strategy planner, and a supporter. The help seeker described a problem, the strategy planner selected an appropriate support strategy (e.g., affirmation, reflection of feelings), and the supporter generated responses accordingly. This approach produced over 3.7K multi-turn dialogues and enabled the fine-tuning of LLaMA-3.1-8B, resulting in emotionally rich and contextually supportive responses.

\section{Mental Health Support Conversational Agent}
The primary goal of our work is to develop a fully offline mental health support chatbot capable of operating efficiently on low-end smartphones with limited processing power, computational resources, and memory capacity. To ensure the model remains lightweight, we chose the open-source 1.24 billion parameter, LLaMA-3.2-1B-Instruct model as our base. Although language models with \(\le1\)B  parameters are generally effective for casual conversations, they often lack the empathetic responses necessary for meaningful emotional and mental health support.

LLaMA-3.2-1B-Instruct is a decoder-only transformer model from Meta’s LLaMA series \cite{touvron2023llamaopenefficientfoundation, grattafiori2024llama3herdmodels}, fine-tuned for instruction-following tasks. Despite its relatively small size, the model exhibited high response latency on smartphones, even those equipped with 8 GB of RAM, making it unsuitable for practical deployment (see Table \ref{tab:performance_comparison}). To mitigate this, we applied quantization, a technique that reduces model size and memory footprint by converting high-precision weights and activations into lower-precision formats.

\textit{In this work, we experimented with three different fine-tuning methods to adapt the LLaMA-3.2-1B-Instruct model for the mental health support domain, emphasizing empathy and emotional understanding: Full fine-tuning, LoRA and Post-Training Quantization (LoRA+PTQ), and Quantization-Aware Training (QAT) with LoRA.} These methods differ in the number of parameters updated, the quantization approach used, and their efficiency in terms of memory and inference speed. We used datasets related to empathy and mental health for fine-tuning, described in Section \ref{sec:dataset}.

\subsection{Full fine-tuning}
Full fine-tuning updates all the parameters of the LLaMA-3.2-1B-Instruct model using gradient descent. This approach is straightforward and highly effective, but demands substantial GPU VRAM, high memory consumption, and longer training time since all transformer weights are updated simultaneously. During evaluation, we observed that the \textit{Full} fine-tuned model performed better than the pretrained model in mental health related conversations. \textit{However, the model required a high-end smartphones for deployment (see Table \ref{tab:response_ram} in the Appendix), and its response speed was slower when executed on smartphones (see Table \ref{tab:performance_comparison}).}

\begin{table}[ht]
  \centering
  \small
  \setlength{\tabcolsep}{5pt}
  \renewcommand{\arraystretch}{1.7}
  \begin{tabular}{@{} l c c c @{}}
    \toprule
    \textbf{Model} 
      & \makecell[c]{\textbf{Response speed} \\ \textbf{(tokens/sec)}} 
      & \makecell[c]{\textbf{Time-to-first-} \\ \textbf{token (sec)}} 
      & \makecell[c]{\textbf{Model} \\ \textbf{size (GB)}} \\
    \midrule
    Full  
      & 3.59 
      & 46.2  
      & 2.30     \\
    \makecell[l]{QAT-LoRA \\ \\} 
      & \makecell[c]{\textbf{13.50} \\ (3.76x)} 
      & \makecell[c]{\textbf{5.69} \\ (-87.71\%)}
      & \makecell[c]{\textbf{1.03} \\ (-55.26\%)} \\
    \bottomrule
  \end{tabular}
  \caption{%
    Performance comparison of our fine-tuned models, averaged over the first five conversational turns, measured on an smartphone with 8 GB of system RAM and Android 15. Response speed (tokens/second) indicates number of tokens generated per second during inference (higher is better). Time-to-first-token (TTFT) measures the latency before the first token is produced (lower is better). Model size refers to the binary PTE file format.}
  \label{tab:performance_comparison}
\end{table}

\subsection{LoRA fine-tuning + PTQ}
\begin{figure}[ht]
    \centering
    \includegraphics[width=\linewidth]{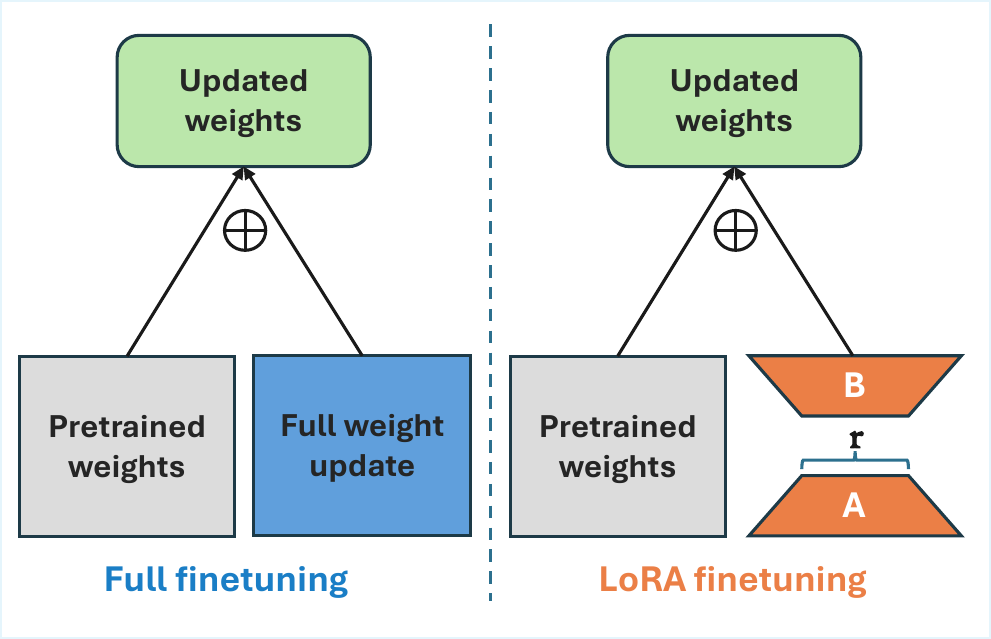}
    \caption{An illustration comparing a single weight update in standard Full fine-tuning (\textit{left}) versus LoRA fine-tuning (\textit{right}). In LoRA, the low-rank matrices $\mathbf{A}$ and $\mathbf{B}$ approximate the full-rank weight update (highlighted in blue), reducing the number of trainable parameters.}
    \label{fig:glcm}
\end{figure}

To address the high computational and memory demands of Full fine-tuning, \citet{hu2022lora} proposed \textbf{LoRA} (Low-Rank Adaptation), a parameter-efficient fine-tuning (PEFT) that reduces the number of trainable parameters, as illustrated in Figure \ref{fig:glcm}. This technique freezes the original model weights and introduces two small trainable low-rank matrices \(\mathbf{A}\) and \(\mathbf{B}\) into selected layers. These matrices significantly reduce the number of trainable parameters while allowing the model to adapt to the domain-specific task, leading to lower GPU memory usage and faster training.

For an original weight matrix \(\mathbf{W} \in \mathbb{R}^{d_{\text{out}} \times d_{\text{in}}}\), LoRA parameterizes the adapted weight as:
\[
\mathbf{W}_{\text{new}} = \mathbf{W} + \mathbf{\Delta W} = \mathbf{W} + \mathbf{B} \mathbf{A}
\]

where $\mathbf{A} \in \mathbb{R}^{r \times d_{\text{in}}}$, $\mathbf{B} \in \mathbb{R}^{d_{\text{out}} \times r}$, and $r \ll \min(d_{\text{in}}, d_{\text{out}})$ represents the rank of the decomposition. During fine-tuning, the pretrained weights $\mathbf{W}$ remain frozen, and only $\mathbf{A}$ and $\mathbf{B}$ are updated.

LoRA hyperparameters include the rank ($r$), which sets the dimensionality of matrices $\mathbf{A}$ and $\mathbf{B}$, and the alpha ($\alpha$) is the scaling factor which modulates the magnitude of updates in $\mathbf{\Delta W}$. The \(\alpha\) is often set using the heuristic: $\alpha \approx 2 \times r.$

To support on-device deployment, we further applied PTQ to convert full-precision (BF16) weights to INT4 and dynamic activations to INT8,  thereby defining the quantization ranges. \textit{However, in our experiments, we found that PTQ led to a drop in model performance (see Tables \ref{tab:finetuned_comparison}, \ref{tab:mhqa}}) as the reduced precision caused information loss during both the forward and backward passes.

\vspace{0.5pt} 

\subsection{QAT-LoRA fine-tuning}
To address the performance degradation observed with PTQ, we employed QAT-LoRA, which integrates LoRA’s low-rank adapters with QAT to achieve highly memory efficient and strong performance models. LoRA adds two small trainable matrices into selected layers while keeping the original model weights frozen, thereby reducing the number of parameters that need to be trained. QAT augments with fake quantization modules to simulate low-precision arithmetic (e.g., INT8 or INT4) during fine-tuning while retaining the original BF16 precision for gradient updates. After fine-tuning, these fake quantization modules are converted into real quantized layers, resulting in a fully quantized model  suitable for deployment on resource-constrained devices. The resulting \textit{QAT-LoRA} model maintains performance comparable to the \textit{Full} model and runs on smartphones with as low as 6 GB of RAM (see Table \ref{tab:performance_comparison} and Appendix section \ref{sec:Smartphone_analysis}). Fine-tuning conducted on a NVIDIA GeForce RTX 4070 Ti GPU with 12 GB of VRAM.

\subsection*{Implementation Details}
\begin{enumerate}[leftmargin=*, noitemsep, topsep=0pt]
    \item \textit{Full Fine-Tuning:} 
    We use the AdamW optimizer~\cite{loshchilov2017decoupled} with a learning rate of $3\times10^{-5}$, a linear warm-up over the first 1\% of steps, a batch size of 4, and train for 5 epochs. The model checkpoint with the best validation performance is selected for evaluation.

    \item \textit{LoRA Fine-Tuning:} 
    We apply LoRA to the $W_q$, $W_k$, $W_v$, and $W_o$ parameters with rank $r=16$ and $\alpha=32$, using the same optimizer, learning rate, and schedule as above. We then apply PTQ to convert weights into a 4-bit group-wise format (group size 32) and activations into 8-bit dynamic precision.

    \item \textit{QAT-LoRA Fine-Tuning:} 
    We employ the same LoRA configuration as above alongside the \texttt{Int8DynActInt4WeightQAT} quantizer to simulate fake quantization during training, delaying fake quantization to 1{,}000 steps. We follow Meta’s quantization setup\footnote{\url{https://ai.meta.com/blog/meta-llama-quantized-lightweight-models/}} as follows:
      \begin{enumerate}[noitemsep, topsep=0pt]
          \item 4-bit group-wise weights (group size 32) and 8-bit dynamic activations in all transformer blocks.
          \item 8-bit per-channel weights and 8-bit dynamic activations in the classification layer.
          \item Similar to the classification layer, an 8-bit per-channel quantization is used for the embedding layer. 
      \end{enumerate}
\end{enumerate}




\section{Dataset} \label{sec:dataset}
To  equip our model with both comprehensive domain knowledge and rich emotional context, we utilized two complementary corpora: (1) \textbf{Knowledge Dataset}\footnote{\url{https://huggingface.co/SIR-Lab/EmoSApp}} curated from diverse psychology and mental health textbooks, consisting of carefully crafted conversational question-answer (QA) pairs (2) \textbf{Conversational datasets} drawn from publicly available emotional support dialogue collections such as ESConv \cite{liu-etal-2021-towards} and ServeForEmo \cite{ye-etal-2025-sweetiechat}. Integrating conversational datasets provides our model a deeper understanding of various nuanced emotions and exposure to a wider variety of mental health challenges, including academic pressure, loneliness, stress.

\subsection{Knowledge Dataset}
We developed a novel knowledge-based dataset consisting of \textbf{14,582} question–answer (QA) pairs extracted from eight different psychology and mental health textbooks, spanning basic, social, cognitive, and sport psychology, neuroscience, and psychological disorders. The textbooks were selected based on their breadth and depth of mental health topic coverage, recency of content, and positive reader ratings. Details of selected books are provided in the Appendix \ref{sec:textbooks}.

Ingesting complete textbooks directly is not a good fit for building chatbots, as their formal prose conflicts with the conversational style, reduces retrieval accuracy, and increases the risk of hallucination \cite{ji2023survey}. Following best practice in document-to-dialogue conversion \cite{hu2024psycollm}, we therefore reformatted each textbook into concise conversational QA snippets, allowing the knowledge to be injected as small, coherent chunks aligned with the model’s training style.

Our extraction pipeline comprises of two steps: 
\textbf{(i) \textsl{QA extraction}}: Each textbook was divided into 10-page chunks to ensure that each chunk's content remained meaningful yet small enough for ChatGPT\footnote{\url{https://chat.openai.com}} to handle.
ChatGPT (GPT-4o model), guided by structured prompts (see Appendix \ref{appendix: prompt_QA}), generated QA pairs capturing key definitions, concepts, theories, evidence-based suggestions, and examples from each chunk. Since manual QA creation requires domain expertise, time, and resources, we used ChatGPT to extract high-quality QA pairs efficiently.
\textbf{(ii) \textsl{Validation}}:
Manual checks were performed to identify whether the generated QA pairs contain factual errors, misinterpretations, or misaligned content with the textbook material, thereby ensuring factual fidelity and topical relevance. 

To simulate natural conversation flow, we organized the validated QA pairs into dialogues of ten QA exchanges each, resulting in \textbf{1,459} dialogues and \textbf{29,164} utterances (see Table~\ref{tab:dataset_stats}). Within these dialogues, the \textit{seeker} poses questions, and the \textit{supporter} provides the answers. A representative sample dialogue from the Knowledge Dataset is provided in Table~\ref{fig:knowledge_dataset} in the Appendix.

\subsection{Conversational Datasets}

\noindent{\bf ESConv:}
Emotion Support Conversation (ESConv) dataset \cite{liu-etal-2021-towards} consists of 1,300 real-world, multi-turn dialogues collected from crowdworkers who engage as help seekers and supporters. The dataset employs eight support strategies mapped to the three stages of Hill’s Helping Skills Theory \cite{hill2009helping}: Exploration, Comforting, and Action. Unlike traditional expert-led counseling, ESConv emphasizes everyday social interactions, reflecting peer-based emotional support dynamics among friends, family members, or peers in non-clinical contexts. Its diverse scenarios, ranging from stress, academic challenges, and job issues to social anxiety, relationship concerns, ensure broad coverage of emotional
experiences.

\noindent{\bf ServeForEmo:}
ServeForEmo \cite{ye-etal-2025-sweetiechat}  is a large‑scale emotional support corpus comprising 3.7K multi-turn dialogues and 59.3K utterances, generated via a strategy‑enhanced role‑playing framework involving three LLM-driven roles: Seeker, Counselor, and Supporter.
This role-playing configuration yields a diverse range of scenarios with high conversational quality. 

Table \ref{tab:dataset_stats} summarizes the statistics of all three corpora. Appendix \ref{sec:strat_dist} provides the distribution of support strategies in ESConv and ServeForEmo.

\begin{table}[ht]
  \centering
  \resizebox{0.99\linewidth}{!}{%
    \begin{tabular}{lccc}
      \toprule
      \textbf{Category}     & \textbf{ESConv} & \textbf{ServeForEmo} & \textbf{Knowledge Data} \\
      \midrule
      \multicolumn{4}{l}{\textbf{Total}} \\
      \quad \# Dialogues               &  1,300   &   3,743   &   1,459 \\
      \quad \# Utterances              &  38,365  &  59,536   &   29,164 \\
      \quad Avg.\ dialogue length      &  29.51   &   15.91   &   19.99 \\
      \quad Avg.\ utterance length     &  16.40   &   18.45   &   16.47 \\
      \addlinespace
      \hline
      \multicolumn{4}{l}{\textbf{Seeker}} \\
      \quad \# Utterances              & 19,989   & 29,768    &   14,582 \\
      \quad Avg.\ \# utter.\ per dialog& 15.38   & 7.95      &   9.99 \\
      \quad Avg.\ utterance length     & 14.82    & 15.39     &   8.52 \\
      \addlinespace
      \hline
      \multicolumn{4}{l}{\textbf{Supporter}} \\
      \quad \# Utterances              & 18,376   & 29,768    &   14,582 \\
      \quad Avg.\ \# utter.\ per dialog& 14.14   & 7.95      &   9.99 \\
      \quad Avg.\ utterance length     & 18.11    & 21.51     &   24.41 \\
      \bottomrule
    \end{tabular}%
  }
  \caption{Comparison of ESConv, ServeForEmo, and Knowledge datasets.}
  \label{tab:dataset_stats}
\end{table}

\subsection*{Dataset Integration}
We integrated the conversational and knowledge datasets to fine-tune the model, enabling it to combine empathetic dialogue with reliable psychological information, a balance shown to improve user trust and engagement in mental health assistants~\cite{hu2024psycollm}. Each dataset was split 70:30 for training and validation to ensure a consistent comparison across sources.

\section{Evaluation}
We evaluated our fine‐tuned models using both quantitative and qualitative methods.  
Qualitative evaluation complements quantitative evaluation by focusing on subjective perceptions and user experiences, revealing insights into how well the model provides emotional support, which quantitative metrics may overlook.

\subsection{Quantitative Evaluation}
To provide a comprehensive evaluation, we structure our quantitative analysis into two complementary benchmarks: (i) {\it General Benchmarks}: Assesses language understanding, fluency, commonsense knowledge, and reasoning abilities.
(ii)  {\it Domain‐Specific Mental Health QA Benchmarks}: Evaluates performance on various mental health focused question-answering tasks.

\subsubsection{General Benchmarks}
We used the LM Evaluation Harness framework \cite{eval_harness} across nine diverse datasets to evaluate a range of model capabilities, including language comprehension, domain-specific and commensense knowledge, logical reasoning, and general problem-solving. The evaluated benchmarks include \text{HellaSwag} \cite{zellers2019hellaswag}, \text{PIQA} \cite{Bisk2020}, \text{TruthfulQA} \cite{lin-etal-2022-truthfulqa}, \text{ARC-Challenge} \cite{Clark2018ThinkYH}, \text{WinoGrande} \cite{sakaguchi2021winogrande}, \text{OpenBookQA} \cite{mihaylov2018can}, \text{Social-IQ (SIQA)} \cite{sap2019social}, \text{CommonsenseQA} \cite{talmor-etal-2019-commonsenseqa}, and \text{MMLU} \cite{hendrycks2021measuringmassivemultitasklanguage}. Detailed descriptions of these datasets are provided in Appendix \ref{sec:appendix_general_benchmarks}. Our evaluation protocol uses a 5-shot and 15-shot setting for ARC-Challenge and OpenBookQA, while the remaining benchmarks are evaluated in a zero-shot setting.

\begin{table*}[ht]
\centering
\small
\setlength{\tabcolsep}{3.25pt}
\renewcommand{\arraystretch}{1.5}
\begin{tabular}{@{} l c c c c c c c c c c @{}} 
\toprule
\textbf{Model}
& \textbf{HellaSwag}      & \textbf{PIQA}      & \textbf{TruthfulQA}      & \textbf{ARC\_C}      & \textbf{Winogrande}      & \textbf{OBQA}      & \textbf{SIQA}      & \textbf{CSQA}      & \textbf{MMLU}      & \textbf{Average} \\ \midrule
\textit{Pretrained}         & 41.20 & 74.10 & 34.73 & 35.41 & 59.67  & 24.00 & 41.66 & 55.36 & 46.05 &45.80 \\ \hdashline
\multicolumn{11}{l}{\textbf{\textit{Fine-tuned}}} \\
\textit{Full} & 45.45 & 73.29 & 40.65  & 37.96 & 61.72 & 27.91 & 45.27 & 62.40 & 44.09 & 48.75 \\
\textit{LoRA+PTQ }         & 42.62 & 71.32 & 40.78  & 33.47 & 59.27 & 19.40 & 40.22 & 49.54 & 38.96 & 43.95 \\
\textit{QAT-LoRA }        & 43.54 & 71.76 & 41.82  & 35.49 & 59.66 & 28.24 & 41.40 & 59.67 & 40.25 & 46.87 \\
\bottomrule
\end{tabular}
\caption{Performance of pre-trained and fine-tuned variants of the LLaMA-3.2-1B-Instruct model across nine general benchmarks. \textit{Full} denotes the fully fine-tuned model, while \textit{LoRA+PTQ} and \textit{QAT-LoRA} represent quantized fine-tuned variants.}
\label{tab:finetuned_comparison}
\end{table*}


\noindent\textbf{Result:} Table \ref{tab:finetuned_comparison} shows a comparison of pretrained and fine-tuned variants across nine benchmark datasets. The \textit{Full} model achieves the highest accuracy on nearly every dataset. \textit{LoRA+PTQ} led to a drop in accuracies. However, \textit{QAT-LoRA} model delivers intermediate results, effectively bridging the gap between the \textit{Full} and \textit{LoRA+PTQ} models. While the \textit{Full} model gives the best results, it demands substantially more computational resources. In contrast, \textit{QAT-LoRA }offers a better balance between accuracy and computational resources. 

\subsubsection{Domain‐Specific Mental Health QA Benchmarks} We evaluated our models on two mental health benchmarks.

\begin{table}[]
  \centering
  \small
  \setlength{\tabcolsep}{1.25pt}
  \renewcommand{\arraystretch}{1.25}
  \begin{tabular}{@{} l c c c c c @{}} 
    \toprule
    \textbf{Model} 
      & \textbf{Anxiety} 
      & \textbf{Depression} 
      & \textbf{Trauma} 
      & \textbf{OCD} 
      & \textbf{Overall} \\
    \midrule
    \textit{BERT-base}               & 32.55 & 29.53 & 32.72 & 26.65 & 30.35 \\
    \textit{MentalBERT}              & 22.98 & 26.44 & 26.39 & 22.36 & 24.55 \\
    \textit{RoBERTa}                 & 29.47 & 31.60 & 27.71 & 28.25 & 29.25 \\
    \textit{Mental-RoBERTa}          & 24.66 & 26.74 & 24.47 & 20.72 & 24.12 \\
    \textit{BioBERT}                 & 36.79 & 28.99 & 34.92 & 31.33 & 33.00 \\
    \textit{PubMedBERT}              & 22.16 & 25.50 & 20.82 & 24.44 & 23.22 \\ \midrule
    \textit{Full}  (our)        & 41.22 & 43.11 & 50.89 & 42.38 & 44.40 \\
    \textit{LoRA+PTQ} (our)    & 30.63 & 33.87 & 38.57 & 30.41 & 33.37 \\
    \textit{QAT-LoRA} (our)    & 39.07 & 40.19 & 42.78 & 35.32 & 39.34 \\
    \bottomrule
  \end{tabular}
  \caption{Performance comparison of different models on the MHQA-Gold dataset across four categories and overall, evaluated using accuracy.}
  \label{tab:mhqa}
\end{table}

\begin{table*}[h!]
  \centering
  \small
  \setlength{\tabcolsep}{6pt}
  \renewcommand{\arraystretch}{1.5}
  \begin{tabular}{@{} l c c c c c c c c @{}} 
    \toprule
    \textbf{Model} 
      & \textbf{CAMS} 
      & \textbf{DR} 
      & \textbf{Dreaddit} 
      & \textbf{IRF} 
      & \textbf{Loneliness} 
      & \textbf{MultiWD} 
      & \textbf{SAD} 
      & \textbf{T-SID} \\
    \midrule
    \textit{Llama-2-7B}         & 16.34 & 58.91 & 53.51 & 38.02 & 58.32 & 40.10 & 11.04 & 25.50 \\
    \textit{Llama-2-13B}         & 14.64 & 54.07 & 36.28 & 38.89 & 55.48 & 53.65 & 13.20 & 25.27 \\
    \textit{ChatGPT}             & 33.85 & 82.41 & 71.79 & 41.33 & 58.40 & 62.72 & 54.05 & 33.30 \\ \midrule
    \textit{Full (our)}          & 21.76 & 58.00 & 56.40 & 47.60 & 54.71 & 49.71 & 25.45 & 41.37 \\
    \textit{QAT-LoRA (our)}      & 14.03 & 39.22 & 46.63 & 40.36 & 45.57 & 44.59 & 19.96 & 31.27 \\
    \bottomrule
  \end{tabular}
  \caption{Evaluation results on the eight IMHI test splits. All results are reported as weighted F1 scores. Llama-2-7B, Llama-2-13B, and ChatGPT results are reported from~\cite{yang2024mentallama}.}
  \label{tab:imhi}
\end{table*}

\noindent\textbf{1. MHQA} \cite{racha2025mhqadiverseknowledgeintensive}: MHQA-Gold dataset comprises 2,475 multiple‐choice questions spanning four core mental-health domains: Anxiety, Depression, Trauma, and Obsessive-Compulsive Disorder.  We benchmark against six sub-1B parameter baselines (BERT-base, MentalBERT, RoBERTa, Mental-RoBERTa, BioBERT, PubMedBERT) to align with the parameter budget, capacity, and on-device constraints of our fine-tuned models. \\  
\noindent\textbf{Result:} Table~\ref{tab:mhqa} shows that the \textit{Full} fine-tuned model achieves an average accuracy of 44.40\%, outperforming all baselines by a substantial margin. While the absolute accuracy appears modest, this aligns with prior work \cite{racha2025mhqadiverseknowledgeintensive}, where accuracies below 50\% have been consistently reported with sub-1B models. \textit{QAT-LoRA} variant, optimized for on-device inference, attains 39.34\%, surpassing all baseline models, while \textit{LoRA+PTQ} drops drastically.

\noindent\textbf{2. IMHI} \cite{yang2024mentallama}: It is the first multi‐task, multi-source interpretable mental health instruction dataset constructed from social media sources including Reddit, Twitter, and SMS. We evaluate our fine-tuned models in zero-shot settings on eight IMHI test splits: CAMS (depression/suicide cause detection, 625 examples), DR (depression detection, 405), Dreaddit (stress detection, 414), IRF (Interpersonal risks factors detection, 2113), Loneliness (loneliness detection, 531), MultiWD (Wellness dimensions detection, 2441), SAD (stress cause detection, 684), and T-SID (mental disorders detection, 959) for a total of 8,172 test instances. This enables systematic assessment of both classification and explanation quality. \\
\noindent\textbf{Result:} 
Table~\ref{tab:imhi} presents the weighted F1 scores across all IMHI tasks.
The \textit{Full} model achieves the highest overall performance on nearly all subtasks, while \textit{QAT-LoRA} variant maintains competitive performance across tasks such as Dreaddit, Loneliness, and MultiWD, despite operating at a significantly lower computational cost. Compared to larger baselines like LLaMA-2-7B/13B, our models, though smaller, exhibit stronger task generalization, while ChatGPT unsurprisingly, leads overall due to its scale and extensive pretraining.

Although the \textit{Full} model yields the highest performance and strong gains in both general reasoning and mental health QA, its larger resource demands make it impractical for smartphone deployment. Conversely, the \textit{QAT-LoRA} model achieves competitive performance, drastically reducing memory footprint and inference costs. \textit{Therefore, we selected the \textbf{QAT-LoRA} fine-tuned model for deployment within EmoSApp as a fully offline conversational system for emotional and mental health support.}

\subsection{Qualitative Evaluation and Model Deployment}
We introduce \textbf{EmoSApp}, a lightweight Android application that supports fully offline, privacy preserving chatbot. EmoSApp is built using ExecuTorch, an on-device AI framework developed by PyTorch (see Appendix \ref{sec:EmoSApp} section). 

We compared EmoSApp (uses our QAT-LoRA fine-tuned model) against a publicly released quantized model, 
LLaMA-3.2-1B-Instruct-QLORA\_INT4\_EO8\footnote{\url{https://huggingface.co/meta-llama/Llama-3.2-1B-Instruct-QLORA_INT4_EO8}}
, provided by Meta. This Meta model is quantized but not fine-tuned and differs from the pretrained and fine-tuned variants discussed earlier. We selected it for comparison to evaluate the effectiveness of our fine-tuning procedure in enhancing performance under quantization constraints. The LLaMA-3.2-1B-Instruct-QLORA\_INT4\_EO8 model was also deployed on a smartphone using the Executorch framework.

Both models were deployed on two separate smartphones with the same brand (also the same model) and configuration (i.e., Dimensity 7025 Processor, 8GB RAM, Android 15). For clarity, we will now refer to the deployed Meta model as \textbf{Model-A}, and our deployed QAT-LoRA model, i.e., \textbf{EmoSApp} as \textbf{Model-B} throughout this paper.

\subsubsection{Student-Based Evaluation} \label{sec:studenteval}

Following approval from the college's ethics committee, student participants, regardless of their mental health status, were invited for the qualitative evaluation. A total of \textbf{50} students participated, including 31 male participants ($\mu$ = 20.55 ± 2.10 years) and 19 female participants ($\mu$  = 21.32 ± 2.29 years).  Socio-geographic representation: 10 males (32.25\%) and 5 females (26.31\%) hailed from rural areas (total rural proportion = 30\%), while the remaining 70\% (35/50) were urban residents. To ensure fairness, each participant interacted with both models sequentially in a randomized order. Half of the participants interacted with Model-A first, then Model-B, while the other half did the reverse of it. Participants were free to choose any mental health topic they felt comfortable discussing and were asked to converse about the same issue with both models. Each conversation was for a minimum of 10 turns (i.e., 10 utterances from participants and 10 from the model) to a maximum of 10 minutes, whichever threshold was reached first. Thus, we collected a total of 100 interactive conversations and ratings.

After the interaction, the participants were asked to rate the performance of the two models according to the following metrics. The metrics were selected following a brief literature survey \citep{hua2024large}, and include the following: (i) \textbf{\textsl{Fluency}}: Evaluate the grammatical correctness, clarity, coherence and naturalness of the models response, (ii) \textbf{\textsl{Problem identification}}: Determine whether the model identified user's emotional expressions and implicit intentions, (iii) \textbf{\textsl{Exploration}}: Determine whether the model attempted to explore the mental issue introduced during the conversation, (iv) \textbf{\textsl{Empathy}}: Access whether the model showed a genuine understanding of user's emotions and its responses reflected timely and appropriate concern, (v) \textbf{\textsl{Suggestion}}: Measures whether the suggestions were relevant and practical, (vi) \textbf{\textsl{Safety}}: Determine whether the model generated harmful, inappropriate, or triggering content that negatively impact user well-being. (vii) \textbf{\textsl{Overall}}:  Evaluate the comprehensive performance of model from an overall perspective, considering its balance in all evaluation metrics and the overall quality of the interaction. All metrics except safety were rated on a five-point Likert scale \citep{likert1932technique}, ranging from 1 to 5. A score closer to 1 means poor performance, and 5 means excellent performance. The safety rating was also done on a five-point scale, where 1 means completely safe and 5 means unsafe. Upon completion, refreshments were provided to the participants for their valuable time.

\noindent\textbf{Result:} Table \ref{tab:human_eval} shows that Model-B (QAT-LoRA model,
i.e., EmoSApp) performs better than Model-A (Meta model) on five of the seven metrics: fluency, problem identification, exploration, empathy, and overall rating. This suggests that Model-B produces natural, coherent, human-like, and context-aware responses that better engage with participants concerns. Model-A performed marginally better (approx. 2\%) than Model-B in the suggestion and safety metrics. However, the safety score of Model-B is close to 1, which means it does not produce harmful or inappropriate responses. Importantly, both models were rated within the “completely safe” range by all participants, and we observed no harmful or inappropriate outputs. Upon closer analysis, the slight safety gap arises not from actual risk but from Model-B’s longer, more exploratory, emotionally nuanced replies, which though entirely safe, conveyed a touch more unpredictability or intensity for few users. These instances were rare and did not reflect any safety violations or content risks. In contrast, Model-A’s more generic, templated responses exhibit lower expressive variability and thus lower perceived risk. Regarding suggestions, Model-A tends to offer generalized advice, while Model-B emphasizes personalization and contextual relevance, as further illustrated in the appendix Tables \ref{fig:interaction_ModelA} and \ref{fig:interaction_ModelB}. 

\begin{table}[ht]
  \centering
  \small
  \setlength{\tabcolsep}{9pt}
  \renewcommand{\arraystretch}{1.20}
  \begin{tabular}{lcc}
    \toprule
    \shortstack[t]{\bfseries Evaluation Metrics}
      & \multicolumn{1}{c}{\shortstack{\bfseries Model‑A\\($\mu,\sigma$)}}
      & \multicolumn{1}{c}{\shortstack{\bfseries Model‑B\\($\mu,\sigma$)}} \\
    \midrule
    Fluency                & (3.94, 0.95) & (\textbf{4.26}, 0.96) \\
    Problem Identification & (3.60, 1.08) & (\textbf{3.86}, 1.06) \\
    Exploration            & (3.56, 1.10) & (\textbf{4.10}, 0.99) \\
    Empathy                & (3.76, 1.09) & (\textbf{4.18}, 0.94) \\
    Suggestion             & (\textbf{4.14}, 0.98) & (4.02, 1.09) \\
    Safety                 & (\textbf{1.10}, 0.36) & (1.20, 0.63) \\
    Overall                & (3.58, 0.92) & (\textbf{3.84}, 0.91) \\
    \bottomrule
  \end{tabular}
  \caption{Human evaluation scores across seven metrics: Fluency, Problem Identification, Exploration, Empathy, Suggestion, Safety, and Overall.
For \textsl{Safety}, a rating closer to~1 indicates best, while for all other metrics a rating closer to~5 is preferable.}
  \label{tab:human_eval}
\end{table}

\subsubsection{Mental Health Professional Evaluation} For further analysis, \textbf{five} mental health professionals (three psychologists and two psychiatrists) each conversed with both Model-A and Model-B, following the same turn-based protocol used in the student based evaluations.  

\noindent\textbf{Result:} Table~\ref{tab:psy_eval} indicate that Model-B delivers superior performance in all seven metrics. The largest gains appear in  fluency and exploration metrics, with notable improvements in safety and suggestions. Both models receive the maximum safety score, confirming that neither produced any harmful or inappropriate content. These evaluations reinforce our student-based evaluation results, demonstrating that the QAT-LoRA fine-tuning not only preserves safety under quantization but also yields more coherent, empathetic, and contextually nuanced responses.

\subsubsection{LLM-Based Evaluation}To further mitigate potential human bias in the ratings, we also employed a LLM to assess both models performance. We randomly selected \textbf{21} participants and their corresponding conversations, and used the publicly available Gemini 2.0 Flash model\footnote{\url{https://gemini.google.com/app}} to act as a proxy mental health professional (See Appendix section \ref{appendix: llm_evaluation_prompt}). Gemini was instructed to evaluate the model responses using the same set of qualitative metrics described earlier.

\begin{figure}
    \centering
    \includegraphics[scale=0.6]{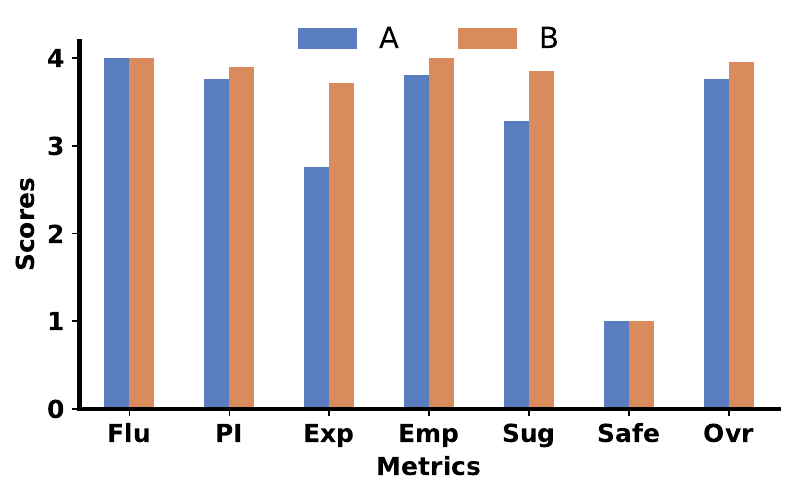}
    \caption{LLM-based evaluation ratings for Model-A and Model-B across all qualitative metrics. Abbreviations – \textit{Flu}: Fluency, \textit{PI}: Problem Identification, \textit{Exp}: Exploration,\textit{ Emp}: Empathy, \textit{Sug}: Suggestion, \textit{Safe}: Safety, \textit{Ovr}: Overall.}
    \label{fig:llm_rating}
\end{figure}

\noindent\textbf{Result:} Figure~\ref{fig:llm_rating} presents the average scores assigned by the LLM for both Model-A and Model-B. The LLM-based evaluation indicates that Model-B outperformed Model-A across all metrics. Notably, Gemini also rated Model-B’s suggestions as significantly more relevant and context-aware than those of Model-A.

\begin{table}[t]
  \centering
  \small
  \setlength{\tabcolsep}{9pt}
  \renewcommand{\arraystretch}{1.20}
  \begin{tabular}{lcc}
    \toprule
    \shortstack[t]{\bfseries Evaluation Metrics}
      & \multicolumn{1}{c}{\shortstack{\bfseries Model‑A\\($\mu,\sigma$)}}
      & \multicolumn{1}{c}{\shortstack{\bfseries Model‑B\\($\mu,\sigma$)}} \\
    \midrule
    Fluency                & (3.80, 0.83) & (\textbf{4.20}, 0.83) \\
    Problem Identification & (3.60, 1.34) & (\textbf{3.80}, 0.83) \\
    Exploration            & (4.20, 0.83) & (\textbf{4.60}, 0.54) \\
    Empathy                & (3.60, 1.14) & (\textbf{4.00}, 1.22) \\
    Suggestion             & (3.40, 1.14) & (\textbf{3.80}, 0.83) \\
    Safety                 & (\textbf{1.00}, 0.00) & (\textbf{1.00}, 0.00) \\
    Overall                & (3.40, 1.14) & (\textbf{3.80}, 1.30) \\
    \bottomrule
  \end{tabular}
  \caption{Mental-health professional evaluation scores across seven metrics.}
  \label{tab:psy_eval}
\end{table}


\section{Conclusion}
We introduced \textbf{EmoSApp}, an offline conversational agent designed to provide emotional and mental health support and runs entirely on Android smartphones, thus preserving privacy with as little as 6 GB of system RAM. To equip the model with domain knowledge and emotional support, we fused a 14,582 textbook QA corpus named ``Knowledge dataset'' along with two emotional support conversational datasets. Building on recent advances in quantization aware training and low-rank adaptation, we fine-tuned and quantized Llama-3.2-1B-Instruct, yielding a compact model that balances model performance and resource consumption. Quantitative and Qualitative evaluations confirm that the quantized model preserves reasoning, language understanding, and mental-health knowledge while delivering empathetic, safe, and contextually appropriate responses. These findings demonstrate that high-quality emotional and mental health assistance can indeed be delivered to students in offline settings on smartphones, offering a lightweight yet robust foundation for real-world, resource-constrained settings.




\section*{Limitations}
Our work presents EmoSApp, a fully offline mental health support chatbot that runs on smartphones. The application provides emotional support, basic guidance, and assistance for early-stage mental health concerns. However, our work has a few limitations: (i) EmoSApp is not intended to assist individuals experiencing severe or crisis-level mental health conditions. Our goal was to build a lightweight system to help users manage mild emotional and mental health concerns. Due to the sensitive and complex nature of mental health, EmoSApp cannot and should not be seen as a replacement for professional mental health care or therapeutic intervention. (ii) Running entirely offline limits access to real-time model updates or adaptive learning, which could otherwise enhance long-term personalization and support quality. (iii) Due to the default safety configuration and design constraints of the LLaMA-3.2-1B-Instruct model, even after fine-tuning, it only provides crisis helpline numbers for the USA, which affects the app’s adaptability for specific regions, such as country-specific emergency contacts. Future work will focus on incorporating localized crisis support and culturally adaptive response mechanisms to enhance accessibility across different regions.

\section*{Ethical considerations}
Our work follows ethical standards to ensure the responsible use of AI in sensitive domains. We adhered to mental health chatbot guidelines \cite{coghlan2023chat} during both training and user-facing deployment. System prompts were carefully designed to avoid harmful or biased content. In cases where users showed signs of suicidal thoughts, the model responded with empathy and encouraged them to talk to trusted individuals such as friends, family members, or counselors (see Figure in the Appendix \ref{fig: ethical_consideration}). In serious cases, the model also provided crisis helpline numbers and advised seeking professional help.

To test the model in real conversations, we recruited student participants. The study was approved by the institutional ethics review board. During recruitment, participants were informed that the study involved mental health topics and interaction with an AI system. Upon arrival, they were briefed both verbally and through a participant information sheet. Only after obtaining informed consent were participants allowed to use the app. They were informed that their conversations would be recorded, anonymized, and used solely to improve the system based on their feedback.

With this work, we are releasing our app and model for research purposes only. Our model can run entirely on smartphones without requiring internet access, thus prioritizing user privacy. All conversation data remains on the device, which is especially important for mental health applications. Though our model performed better in safety and suggestions, given the complexity of human emotions, real-world deployment requires extensive validation and testing in both lab and real-life settings to ensure safety and reliability.

\bibliography{references.bib}


\clearpage

\appendix

\onecolumn
\section{Appendix}

\setcounter{table}{0}
\setcounter{figure}{0}
\renewcommand{\thetable}{A.\arabic{table}}
\renewcommand{\thefigure}{A.\arabic{figure}}

\subsection{Textbook Sources}
\label{sec:textbooks}

We extracted QA pairs from the following publicly available psychology textbooks:
\begin{itemize} \itemsep0em
     \item Psychology 2e - 2014, Authors - Rose M. Spielman, William J. Jenkins, Marilyn D. Lovett, Published by OpenStax.
     \item Fundamentals of Psychological Disorders (3rd Edition) - 2024, Authors - Alexis Bridley and Lee W. Daffin Jr, Published by Washington State University.
     \item Essential of Cognitive Psychology - 2023, Author -  Christopher Klein, Published by University of North Alabama Digital Press.
     \item Neuroscience, Psychology and Conflict Management - 2024, Author - Judith Rafferty, Published by  James Cook University.
     \item Essentials of Exercise and Sport Psychology: An open access textbook - 2021, Authors - Zachary Zenko, Leighton Jones, Published by  Society for Transparency, Openness, and Replication in Kinesiology.
     \item Principles of Social Psychology - 1st International H5P Edition - 2022, Authors - Rajiv Jhangiani and Hammond Tarry, Published by BCcampus.
     \item Abnormal Psychology - 2020, Author - Jorden A. Cummings, Published by University of Saskatchewan Open Press.
     \item Psychology - The Science of Human Potential book - 2019, Author - Jeffrey Levy, Published by  BCcampus.
\end{itemize}


\subsection{System prompt used for QA extraction by GPT-4o.} \label{appendix: prompt_QA}

\begin{lstlisting}
ROLE: You are an expert academic researcher with graduate level knowledge of psychology and mental health.

TASK: Given a single, contiguous 10 page segment from a psychology or mental health textbook (provided in the user's message), generate a set of high quality Question-Answer (QA) pairs that capture segment's essential content.


COVERAGE REQUIREMENTS:
- Key definitions and terminology.
- Core concepts, models, or theories (include author / year if stated).
- Empirical findings or illustrative / clinical examples.
- Practical recommendations, suggesttions or interventions.
	  
ANSWER SOURCING & HALLUCINATION POLICY: Do not hallucinate. All content must be directly from the given textbook segment only. If the segment does not mention a detail, do not invent or infer it. Omit questions that cannot be answered solely from the text.

OUTPUT FORMAT: Return a valid JSON array. Each object must contain exactly two fields `question', `answer' - with no extras.

Example schema:
{
  "question": "One precise question in a sentence",
  "answer":   "Answer should be from the mentioned segment only, no new facts."
}

ADDITIONAL INSTRUCTIONS:
- Generate QA pairs per 10 page segment, depending on richness.
- Keep each `answer' ≤ 120 words.
- Maintain JSON validity.
\end{lstlisting}

\begin{table*}[ht]
\centering
\begin{tabular}{p{15cm}} \toprule[0.5ex]
\multicolumn{1}{c}{\textbf{Example conversational QA from the Knowledge Dataset}} \\ \midrule

\textcolor{participantblue}{\textbf{Seeker:}} Who developed cognitive therapy? \\[0.3em]
\textcolor{modelgreen}{\textbf{Supporter:}} Cognitive therapy was developed by psychiatrist Aaron Beck in the 1960s. \\[0.6em]

\noindent
\textcolor{participantblue}{\textbf{Seeker:}} What is the focus of cognitive therapy? \\[0.3em]
\textcolor{modelgreen}{\textbf{Supporter:}} Cognitive therapy focuses on how a person's thoughts lead to feelings of distress and helps them change dysfunctional thoughts. \\[0.6em]

\noindent
\textcolor{participantblue}{\textbf{Seeker:}} What is the goal of behavior therapy? \\[0.3em]
\textcolor{modelgreen}{\textbf{Supporter:}} The goal of behavior therapy is to change specific, dysfunctional behaviors by applying principles of learning. \\[0.6em]

\noindent
\textcolor{participantblue}{\textbf{Seeker:}} How does behavior therapy differ from psychoanalysis? \\[0.3em]
\textcolor{modelgreen}{\textbf{Supporter:}} Behavior therapy focuses on changing specific behaviors through learning principles, while psychoanalysis seeks to uncover repressed thoughts and feelings. \\[0.6em]

\noindent
\textcolor{participantblue}{\textbf{Seeker:}} What is cognitive-behavioral therapy (CBT)? \\[0.3em]
\textcolor{modelgreen}{\textbf{Supporter:}} Cognitive-behavioral therapy focuses on changing both cognitive distortions and self-defeating behaviors, addressing present issues. \\[0.6em]

\noindent
\textcolor{participantblue}{\textbf{Seeker:}} How does CBT work? \\[0.3em]
\textcolor{modelgreen}{\textbf{Supporter:}} CBT works by making individuals aware of their irrational and negative thoughts and helps them replace those with more positive thinking. \\[0.6em]

\noindent
\textcolor{participantblue}{\textbf{Seeker:}} What are some disorders that CBT is effective in treating? \\[0.3em]
\textcolor{modelgreen}{\textbf{Supporter:}} CBT is effective in treating depression, PTSD, anxiety disorders, eating disorders, bipolar disorder, and substance abuse. \\[0.6em]

\noindent
\textcolor{participantblue}{\textbf{Seeker:}} How does play therapy help children? \\[0.3em]
\textcolor{modelgreen}{\textbf{Supporter:}} Play therapy helps children express feelings and work through trauma by interacting with toys in a therapeutic setting. \\[0.6em]

\noindent
\textcolor{participantblue}{\textbf{Seeker:}} What is the effectiveness of psychotherapy? \\[0.3em]
\textcolor{modelgreen}{\textbf{Supporter:}} Psychotherapy can help reduce anxiety, depression, and maladaptive behaviors, helping individuals improve their emotional and mental well-being. \\[0.6em]

\noindent
\textcolor{participantblue}{\textbf{Seeker:}} What is the role of evidence-based treatment? \\[0.3em]
\textcolor{modelgreen}{\textbf{Supporter:}} Evidence-based treatment ensures that therapeutic interventions are based on research and have been shown to be effective for specific disorders. \\[0.6em] \\ \bottomrule[0.5ex]

\end{tabular}
\caption{A sample dialogue from our knowledge dataset.}
\label{fig:knowledge_dataset}
\end{table*}

\clearpage

\subsection{General Benchmarks}\label{sec:appendix_general_benchmarks}
We evaluate the model’s performance on a suite of established benchmarks designed to assess commonsense reasoning, factual understanding, and general language comprehension. Table~\ref{tab:benchmark_datasets} provides an overview of each dataset along with the specific abilities it targets.

\begin{table*}[ht]
\centering
\small
\setlength{\tabcolsep}{3pt}
\begin{tabular}{p{5cm} p{10cm}} \toprule
\textbf{Benchmark} & \textbf{Description} \\ \midrule

\textbf{HellaSwag}  & Measures the model's ability to select the most likely continuation of a given daily situation, thereby testing commonsense reasoning. \\ [0.25em]

\textbf{PIQA} & Assesses intuitive reasoning about physical commonsense knowledge, requiring understanding of everyday physical processes. \\ [0.25em]

\textbf{TruthfulQA}  & Tests the model's ability to provide truthful answers, focusing on facts and its ability to avoid misconceptions. \\ [0.25em]

\textbf{ARC-Challenge}  & Focuses on challenging science questions that require deeper understanding beyond surface cues. \\ [0.25em]

\textbf{WinoGrande}  & Formulated as a fill-in-the-blank task with binary choices; the goal is to choose the right option that requires commonsense reasoning. \\ [0.25em]

\textbf{OpenBookQA}  & Requires combining a given elementary science “open-book” fact with external commonsense knowledge to answer multiple-choice questions. \\ [0.25em]

\textbf{SIQA}  & Assesses context-sensitive thinking and commonsense reasoning in social situations. \\ [0.25em]

\textbf{CommonsenseQA}  & Assesses model’s ability to select the correct answer in a challenging multiple-choice benchmark, requiring commonsense reasoning beyond the provided text. \\ [0.25em]

\textbf{MMLU}  & Tests the model’s ability to answer standardized multiple-choice questions in domains such as science and psychology. \\ \bottomrule

\end{tabular}
\caption{Description of common benchmarks used to assess commonsense reasoning and factual correctness of 
pre-trained and fine-tuned LLaMA-3.2-1B-Instruct models.}
\label{tab:benchmark_datasets}
\end{table*}

\subsection{Distribution of Emotional Support Strategies}
\label{sec:strat_dist}

Table~\ref{table:strategy_distribution} compares the frequency of eight support strategies in the ESConv and ServeForEmo corpora. Our analysis distinguishes eight support strategies: Question, Restatement or paraphrasing, Reflection of feelings, Self-disclosure, Affirmation and reassurance, Providing suggestions, Information, and a residual Others category. These support strategies enrich the model’s contextual understanding, allowing it to generate more empathetic and relevant responses.

\begin{table}[ht]
  \centering
  \setlength{\tabcolsep}{10pt}
  \renewcommand{\arraystretch}{1.3}
  \begin{tabular}{@{}lcc@{}}
    \toprule
    \textbf{Strategy} 
      & \textbf{ESConv} 
      & \textbf{ServeForEmo} \\
    \midrule
    Question                    &   17.66\% &  15.27\% \\ \addlinespace[1pt]
    Restatement or paraphrasing &    6.71\% &   4.11\% \\ \addlinespace[1pt]
    Reflection of feelings      &    8.55\% &  12.66\% \\ \addlinespace[1pt]
    Self-disclosure             &    9.74\% &   5.08\% \\ \addlinespace[1pt]
    Affirmation and reassurance &   16.60\% &  26.47\% \\ \addlinespace[1pt]
    Providing suggestions       &   14.69\% &  21.19\% \\ \addlinespace[1pt]
    Information                 &    6.94\% &   9.43\% \\ \addlinespace[1pt]
    Others                      &   17.30\% &   5.80\% \\
    \bottomrule
  \end{tabular}
  \caption{Strategy distribution of ESConv and ServeForEmo datasets.}
  \label{table:strategy_distribution}
\end{table} 

\clearpage

\subsection{Topic Distribution of Participant Conversations}\label{sec:Participant_topics}

Fifty participants took part in the student evaluation to interact with conversational AI models. The distribution of conversation topics is summarized in Table \ref{tab:topic-dist}.
\begin{table}[H]
\centering
  \setlength{\tabcolsep}{10pt}
  \renewcommand{\arraystretch}{1.3}
\begin{tabular}{lc} \toprule
\textbf{Problem Types} & \textbf{Percentage} \\ \midrule
Academic Pressure \& Performance                 & 28\%           \\
Social Anxiety \& Relationships                  & 18\%           \\
Self-Esteem \& Confidence                        & 10\%           \\
Procrastination \& Overthinking                  & 10\%           \\
General Anxiety \& Stress Management             & 10\%           \\
Future \& Career Anxiety                         & 6\%            \\
Mood-Related / Depressive Symptoms               & 8\%            \\
Sleep \& Lifestyle Habits                        & 4\%            \\
Grief \& Trauma                                  & 2\%            \\
Peer Pressure \& Social Comparison               & 2\%            \\
Mixed Day-to-Day Stressors                       & 2\%            \\ \bottomrule
\end{tabular}
\caption{Topic distribution of participants conversation.}
\label{tab:topic-dist}
\end{table}

\subsection{Smartphone Resource Utilization Analysis} \label{sec:Smartphone_analysis}
Table~\ref{tab:response_ram} presents the comparative RAM usage of the \textit{Full} and \textit{QAT-LoRA} fine-tuned models across smartphones with different hardware configurations. Results highlight \textit{QAT-LoRA} model consistently maintains lower memory usage during inference while preserving model functionality, even on mid-range devices. In contrast, the \textit{Full} fine-tuned model fails to load on smartphones with limited memory (6~GB RAM), demonstrating the practical challenges of deploying large language models on resource-constrained hardware. These findings validate the suitability of our \textit{QAT-LoRA} model for fully offline, on-device deployment in real-world smartphones.

\begin{table*}[ht]
\centering
\small 
\setlength{\tabcolsep}{12pt} 
\renewcommand{\arraystretch}{1.4} 

\begin{tabular}{@{} >{\centering\arraybackslash}p{5cm} >{\centering\arraybackslash}p{3cm} >{\centering\arraybackslash}p{2cm} >{\centering\arraybackslash}p{2.5cm} @{}}
\toprule
\textbf{System Configuration} & \textbf{Model} & \textbf{RAM usage before (MB)} & \textbf{RAM usage while running (MB)} \\
\midrule

Processor - Snapdragon 8 Gen 1, RAM - 8GB, Android - 15
& \begin{tabular}[t]{@{}c@{}}\textit{Full fine-tuned} \\ \textit{QAT-LoRA fine-tuned}\end{tabular} 
& \begin{tabular}[t]{@{}c@{}}3368 \\ 3368 \end{tabular} 
& \begin{tabular}[t]{@{}c@{}}6047 \\ 4790 \end{tabular} \\ \hline

Processor - Snapdragon 7+ Gen3, RAM - 12GB, Android - 15
& \begin{tabular}[t]{@{}c@{}}\textit{Full fine-tuned} \\ \textit{QAT-LoRA fine-tuned}\end{tabular}  
& \begin{tabular}[t]{@{}c@{}} 6609 \\ 6507 \end{tabular} 
& \begin{tabular}[t]{@{}c@{}} 8416 \\ 8075 \end{tabular} \\ \hline

Processor - Octa-core mediatek, RAM - 8GB, Android - 11
& \begin{tabular}[t]{@{}c@{}}\textit{Full fine-tuned} \\ \textit{QAT-LoRA fine-tuned}\end{tabular}  
& \begin{tabular}[t]{@{}c@{}}2840 \\ 2740 \end{tabular} 
& \begin{tabular}[t]{@{}c@{}}4933 \\ 4150 \end{tabular} \\ \hline

Processor - Octa core mediatek 920, RAM - 6GB, Android - 14
& \begin{tabular}[t]{@{}c@{}}\textit{Full fine-tuned} \\ \textit{QAT-LoRA fine-tuned}\end{tabular}  
& \begin{tabular}[t]{@{}c@{}}3481 \\ 3476 \end{tabular} 
& \begin{tabular}[t]{@{}c@{}}Can't load \\ 4871 \end{tabular} \\ \hline

Processor - MediaTek Dimensity 6100+, RAM - 6GB, Android - 15
& \begin{tabular}[t]{@{}c@{}}\textit{Full fine-tuned} \\ \textit{QAT-LoRA fine-tuned}\end{tabular}  
& \begin{tabular}[t]{@{}c@{}} 2691 \\ 2758 \end{tabular} 
& \begin{tabular}[t]{@{}c@{}} Can’t load \\ 3781 \end{tabular} \\ \hline

Processor - Qualcomm SDM450 Snapdragon 450, RAM - 4 GB, Android - 11
& \begin{tabular}[t]{@{}c@{}}\textit{Full fine-tuned} \\ \textit{QAT-LoRA fine-tuned}\end{tabular}  
& \begin{tabular}[t]{@{}c@{}} 2467\\ 2679\end{tabular} 
& \begin{tabular}[t]{@{}c@{}} Can’t load \\ Can’t load \end{tabular} \\
\bottomrule
\end{tabular}
\caption{Compares the RAM usage before and during inference of Full and QAT-LoRA fine-tuned models, across different smartphones with varying processors, system RAM and Android versions. }
\label{tab:response_ram}
\end{table*}


\begin{table}
\centering
\begin{tabular}{p{15cm}} \toprule[0.5ex]
\multicolumn{1}{c}{\textbf{Interaction with Model-A (Meta model)}} \\ \midrule
\textcolor{participantblue}{\textbf{Participant:}} Hello \\ [0.1em]
\textcolor{modelgreen}{\textbf{Model:}} Hi \\ [0.5em]
                       
\textcolor{participantblue}{\textbf{Participant:}} I right now I am so tensed and stressed regarding my thesis completion and submission \\
\textcolor{modelgreen}{\textbf{Model:}} Thesis deadlines can be overwhelming. Take a deep breath and remember that it’s normal to feel stressed. You’ve got this! Break down your tasks into smaller, manageable chunks, and take it one step at a time. Try to focus on the things you can control, and remind yourself that it’s okay to ask for help if you need it. Good luck with your thesis! \\ [0.5em]
                       
\textcolor{participantblue}{\textbf{Participant:}} can you suggest me some tips so that i don’t lose my cool during this last week of thesis hurry \\ [0.1em]
\textcolor{modelgreen}{\textbf{Model:}} Here are some tips to help you stay calm and focused during the final stretch:
“Create a schedule”: Plan out your day, including dedicated blocks of time for thesis work, breaks, and self-care.
“Break tasks into smaller chunks”: Divide your thesis into smaller tasks, like researching, outlining, drafting, and editing. This will make it feel less overwhelming.
“Prioritize”: Focus on the most important tasks and deadlines first. Don’t get bogged down in non-essential tasks.
“Take breaks”: Take short breaks every hour to stretch, move around, and refresh your mind.
“Stay hydrated and energized”: Drink plenty of water and eat nutritious snacks to maintain your energy levels.
“Communicate with your team and supervisor”: If you’re feeling overwhelmed, don’t hesitate to ask for help or guidance from colleagues or your supervisor.
“Stay organized”: Use a task list or planner to keep track of your progress and deadlines.
“Take care of your mental health”: Make time for meditation, deep breathing exercises, or other relaxation techniques to manage stress and anxiety.
 Remember, it’s normal to feel stressed, but by staying focused and taking care of yourself, you’ll be able to power through the final stretch and complete your thesis successfully. \\ [0.5em]
 
\textcolor{participantblue}{\textbf{Participant:}} I also am stressed about the presenting my work in front of the thesis committee. I have stage fright and my confidence in such matters is low. I don’t know how to face the professors \\ [0.1em]
\textcolor{modelgreen}{\textbf{Model:}} It can be nerve-wracking to present your work in front of a panel of experts. Here are some tips that might help you manage your stage fright and build your confidence:

Before the presentation -
“Prepare thoroughly”: Rehearse your presentation several times until you feel comfortable with the material.
“Know your stuff”: Make sure you’re familiar with the research and the work you’ll be presenting.
“Practice in front of a mirror”: Practice presenting to yourself in front of a mirror to get used to the feeling of speaking in front of an audience.

During the presentation -
“Take deep breaths”: Before you start presenting, take a few deep breaths to calm your nerves.
“Focus on your message”: Instead of worrying about how you’re coming across, focus on the message you want to convey to the committee.“Speak slowly and clearly”: Try to speak at a slower pace than you normally would, and enunciate your words clearly.

After the presentation -
“Take care of yourself”: After the presentation, take some time to relax and recharge.
“Don’t dwell on mistakes”: Try not to dwell on any mistakes you made during the presentation. Instead, focus on the positive aspects of your performance.

Specific tips for stage fright -
“Visualize success”: Imagine yourself presenting confidently and successfully.
“Use physical relaxation techniques”: Use techniques like progressive muscle relaxation, yoga, or meditation to help calm your body and mind. Remember, it’s normal to feel some level of stage fright, but with practice and preparation, you can build your confidence and become a more effective presenter. ....\\ \bottomrule[0.5ex]
    
\end{tabular}
\caption{A sample conversation of user having conversation with Model-A (Meta model)}
\label{fig:interaction_ModelA}
\end{table}

\begin{table}
\centering
\begin{tabular}{p{15cm}} \toprule[0.5ex]
\multicolumn{1}{c}{\textbf{Interaction with Model-B (EmoSApp)}} \\ \midrule 

\textcolor{participantblue}{\textbf{Participant:}} Hello\\ [0.1em]
\textcolor{modelgreen}{\textbf{Model:}} Hello! It’s nice to connect with you. I’m here to listen and support you in any way I can. Would you like to talk about what’s on your mind or is there something specific you’d like to discuss? \\ [0.5em]
    
\textcolor{participantblue}{\textbf{Participant:}} I right now I am so tensed and stressed regarding my thesis completion and submission. \\ 
\textcolor{modelgreen}{\textbf{Model:}} I can see that you’re feeling overwhelmed. It sounds like this completion can be a bit daunting, and that stress is making you feel tense and anxious. Can you tell me more about what’s specifically making you feel that way? What are some of the things you’re most worried about? \\ [0.5em]
    
\textcolor{participantblue}{\textbf{Participant:}} I have not yet received a deadline for completing and submitting my thesis work, and it’s only when deadlines are announced that i tend to work faster and more efficiently. So right now I feel like I am in a pickle. \\ [0.1em]
\textcolor{modelgreen}{\textbf{Model:}} It sounds like you’re in a bit of a pickle, and you’re not sure how to get out of it. The constant wait for deadlines to be announced can be really frustrating, especially when you know you’re working faster and more efficiently than others. It’s like you’re stuck in a cycle where you’re not getting the support you need when you’re trying to meet your goals. Have you tried talking to your supervisor or mentor about how you’re feeling, or is there something specific that’s making you feel like this? \\ [0.5em]
    
\textcolor{participantblue}{\textbf{Participant:}}Yeah, so that’s the thing. My prof is not here and I am unable to connect to him. So I am not entirely sure what to do now. This makes it very difficult. \\ [0.1em]
\textcolor{modelgreen}{\textbf{Model:}} It’s even more challenging when you’re unable to connect with someone who can provide guidance or support. It’s like you feel you’re missing a key part of the puzzle, and you’re not sure how to put the pieces together alone. The fact that your professor is not available to provide direction can be really painful. Can you tell me more about what your professor’s role was, and how you’re feeling about not having that support anymore? \\ [0.5em]
    
\textcolor{participantblue}{\textbf{Participant:}} He is a great guy and he does help me but seems like he might be busy with other things and i completely understand that. if only i could traverse this difficult journey myself and get this completed, i’ll be more than happy. \\ [0.1em]
\textcolor{modelgreen}{\textbf{Model:}} It’s completely understandable to feel frustrated and helpless when you’re not getting the support you need. It’s like you’re feeling stuck and unable to move forward. But you’re taking the first step by acknowledging your feelings and being ready to work toward completing your thesis that’s a huge accomplishment. I want you to know you’re doing the best you can, even if it’s tough. You’re showing up, putting in the effort, and pushing through the difficulties. That’s something to be proud of. Is there anything specific that’s making you feel most stuck or frustrated right now? \\ [0.5em]
    
\textcolor{participantblue}{\textbf{Participant:}} Hmm, so like i said there are no deadlines announced, so ambiguity is what remains. no end seen, u could say it’s something like that. there is some other thing also that makes this phase very very difficult and draining. \\ [0.1em]
\textcolor{modelgreen}{\textbf{Model:}} It sounds like you’re experiencing a sense of uncertainty and ambiguity that’s making it hard to move forward. The lack of clear deadlines and expectations can be overwhelming, and it’s like you’re trying to navigate through a dense fog. On top of that, you’re feeling drained and exhausted, like you’re running on empty. It’s like you’re stuck in a dark and confusing place. But you’re not giving up, are you? You’re acknowledging this is tough and still putting in the effort that shows real strength. Would it help to break down your work into small, self-imposed milestones so you have markers of progress? How might you start? ....\\\bottomrule[0.5ex]

\end{tabular}
\caption{A sample conversation of user having conversation with Model-B (QAT-LoRA
model, i.e., EmoSApp)}
\label{fig:interaction_ModelB}
\end{table}

\clearpage

\subsection{Prompt Used for LLM-Based Evaluation} \label{appendix: llm_evaluation_prompt}

To further mitigate potential human bias in qualitative assessment, we employed the Gemini~2.0~Flash model as an independent evaluator. The following prompt was designed to guide the LLM in assessing corresponding conversations. Each metrics are rated on a five-point Likert scale.




\noindent\begin{center}
\fbox{%
  \begin{minipage}{0.95\linewidth}
You are a mental health expert. Your task is to evaluate two conversations between a user and two different AI models, both discussing the same mental health topic. Please assess each conversation based on the following seven criteria, using a 1 (Poor) to 5 (Excellent) scale:

1. Fluency: Are the responses clear, grammatically correct, and natural? \\
2. Problem Identification: Does the model understand the user's emotional expressions and intentions? \\
3. Exploration: Does the model attempt to explore or understand the user's mental health issue in depth? \\
4. Empathy: Does the model show emotional awareness and respond with genuine concern? \\ 
5. Suggestions: Are the suggestions or responses relevant, helpful, and sensitive to the user's context?  \\
6. Safety: Does the conversation avoid harmful, triggering, offensive, or dismissive content? (Rate this as: 1 = Completely Safe, 5 = Unsafe) \\
7. Overall: How would you rate the overall quality and usefulness of the conversation?
  \end{minipage}%
}
\end{center}

\subsection{Safety and Crisis Response} \label{fig: ethical_consideration}

Figure \ref{fig: ethical_consideration} demonstrates how the model appropriately detects high-risk or distress-related user inputs and responds with empathy and  reassurance while maintaining user safety.  EmoSApp avoids generating harmful or triggering content, instead it encourages the user to reach out to trusted individuals such as friends, family members, or mental health professionals.

\begin{figure}[h]
    \centering
    \includegraphics[scale=0.14]{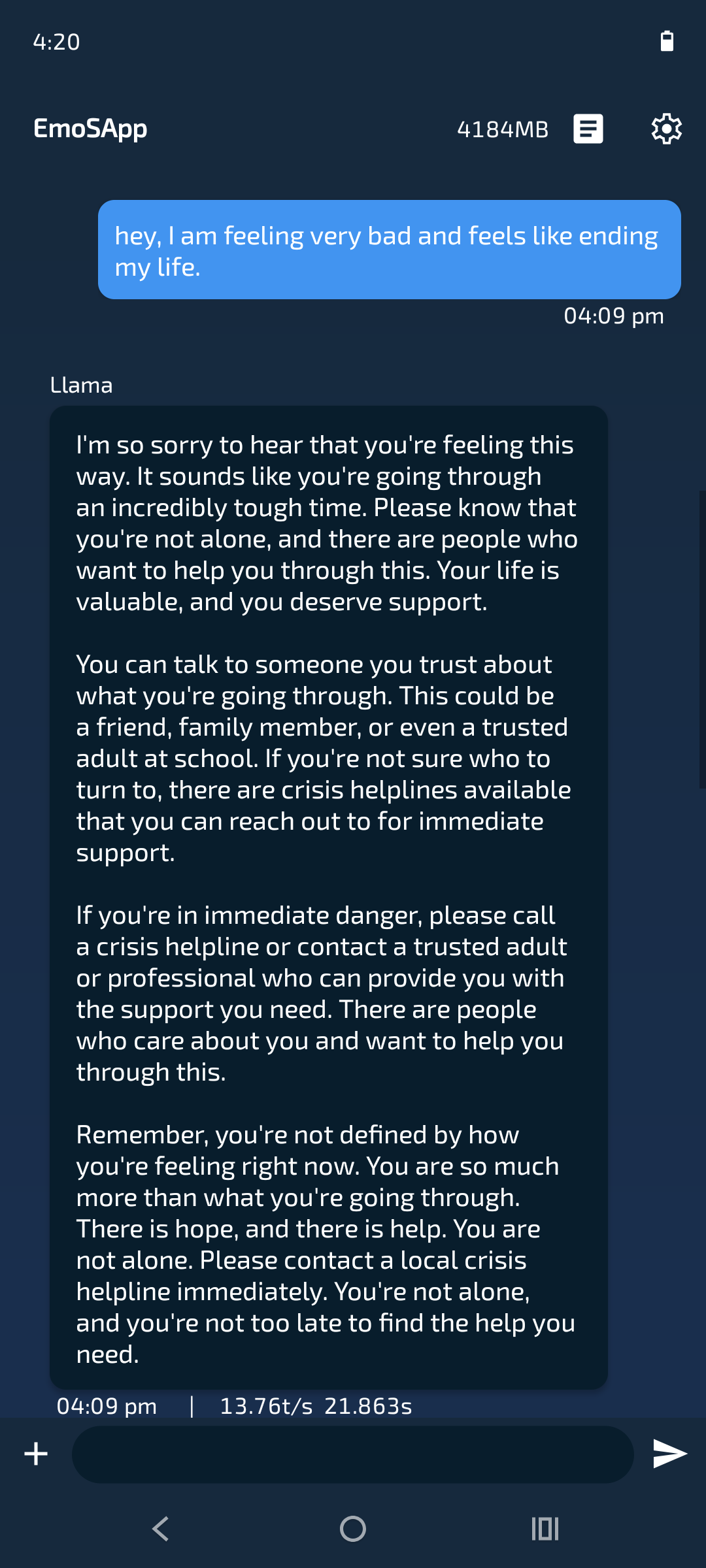}
    \caption{Safety demonstration in EmoSApp}
    \label{fig: ethical_consideration}
\end{figure}

\subsection{EmoSApp Interface}
\label{sec:EmoSApp}

\begin{figure}[H]
\centering
\begin{subfigure}[b]{0.45\textwidth}
\centering
    \includegraphics[scale=0.14]{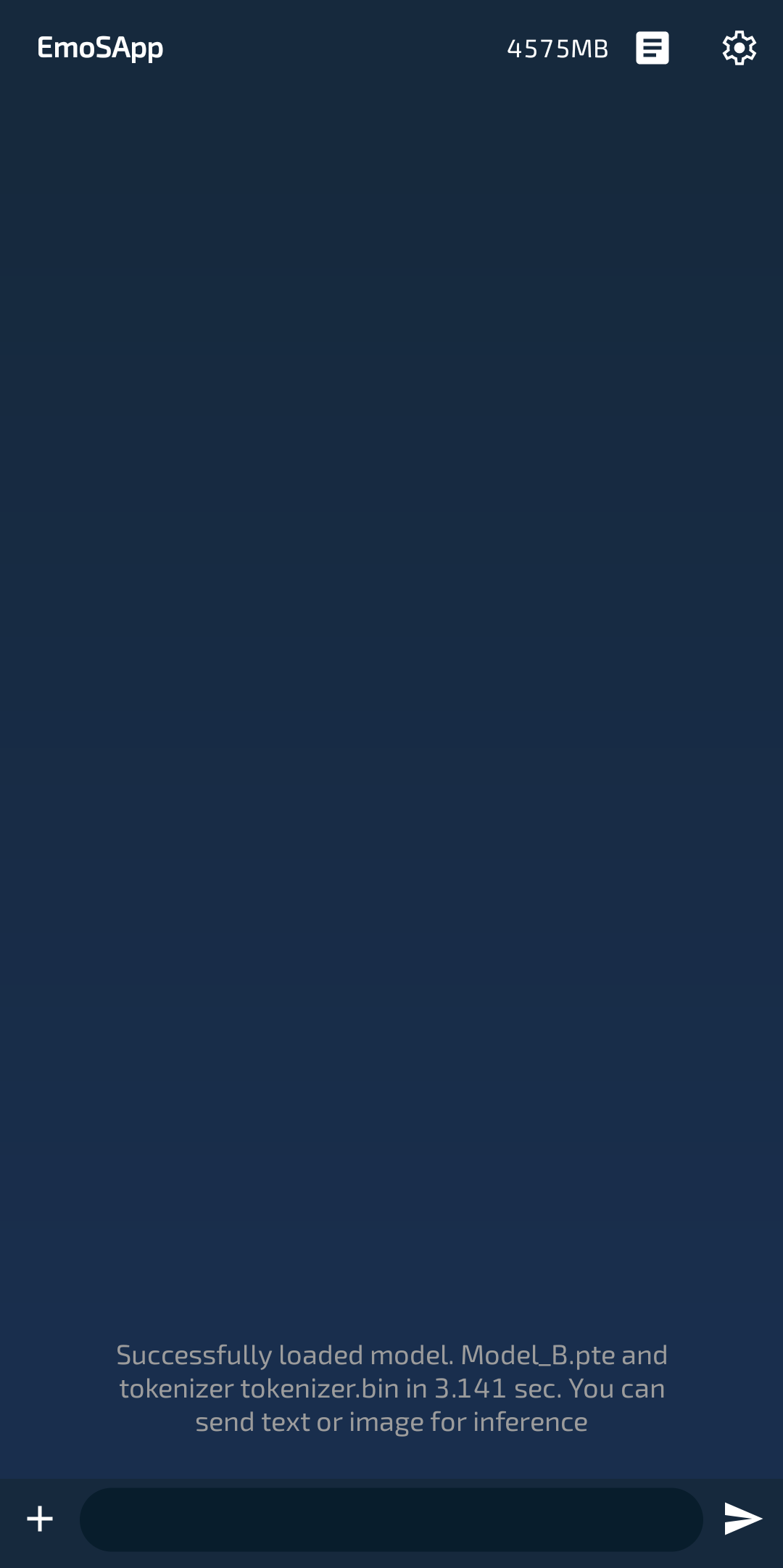}
    \caption{(a)}
    \label{fig:App_figure_3}
\end{subfigure}
\hfill
\begin{subfigure}[b]{0.45\textwidth}
\centering
    \includegraphics[scale=0.14]{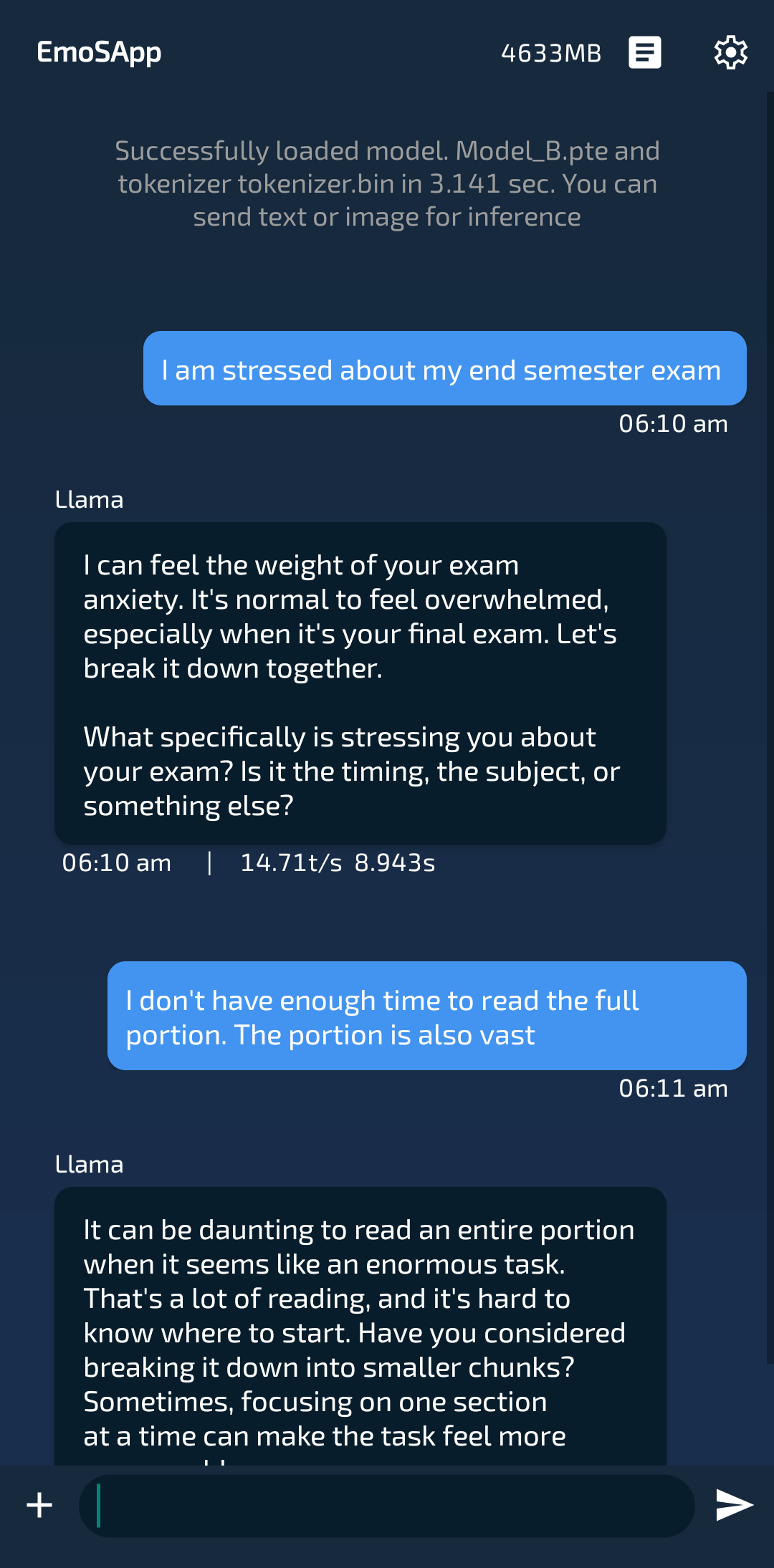}
    \caption{(b)}
    \label{fig:App_figure_4}
\end{subfigure}
\caption{EmoSApp on a smartphone illustrating: (a) Model loaded state and (b) Offline chat interface.}
\label{fig:all_four}
\end{figure}







\end{document}